\documentclass[letterpaper, 10pt, conference]{ieeeconf}

\IEEEoverridecommandlockouts   % needed for \thanks

\usepackage[utf8]{inputenc}
\usepackage[T1]{fontenc}
\usepackage{amsmath,amssymb}
\usepackage{graphicx}
\usepackage{booktabs}
\usepackage{multirow}
\usepackage{array}
\usepackage{tabularx}
\usepackage{cite}
\usepackage[hidelinks]{hyperref}
\usepackage{xcolor}
\newcolumntype{L}[1]{>{\raggedright\arraybackslash}p{#1}}
\newcolumntype{C}{>{\centering\arraybackslash}X}

\makeatletter
\def\paragraph{\@startsection{paragraph}{4}{\z@}%
  {0ex plus 0.1ex minus 0.1ex}%
  {0ex}{\normalfont\normalsize\itshape}}
\makeatother

\makeatletter
\let\WBUMI@defaultmakecaption\@makecaption
\def\fnum@table{TABLE~\thetable}
\long\def\@makecaption#1#2{%
  \ifx\@captype\@IEEEtablestring
    \par\noindent\parbox{\linewidth}{%
      \normalfont\footnotesize
      \sbox\@tempboxa{#1:\enspace #2}%
      \ifdim\wd\@tempboxa>\linewidth
        \noindent #1:\enspace #2\par
      \else
        \noindent\hbox to\linewidth{\hfil\box\@tempboxa\hfil}\par
      \fi}%
    \@IEEEtablecaptionsepspace
  \else
    \WBUMI@defaultmakecaption{#1}{#2}%
  \fi}
\makeatother

\newcommand{\yes}{\textcolor{green!50!black}{\checkmark}}
\newcommand{\no}{\textcolor{red!70!black}{$\times$}}

\newcommand{\nap}{\textcolor{black!45}{--}}
\title{
\textbf{Whole-Body UMI}: Transferring UMI Manipulation Skills to Humanoid
Whole-Body Manipulation via Real-Time Motion Generation}

\author{Yuxuan Nai$^{1,2}$, Leixin Chang$^{1}$, Liangjing Yang$^{1}$, Shuo Yang$^{3}$, Zhongyu Li$^{2,4}$%
\thanks{$^{1}$Zhejiang University;
$^{2}$Hong Kong Embodied AI Lab;
$^{3}$Mondo Robotics;
$^{4}$The Chinese University of Hong Kong.}%
}
\IEEEaftertitletext{%
\begin{minipage}{\textwidth}%
\centering%
\includegraphics[width=1\textwidth]{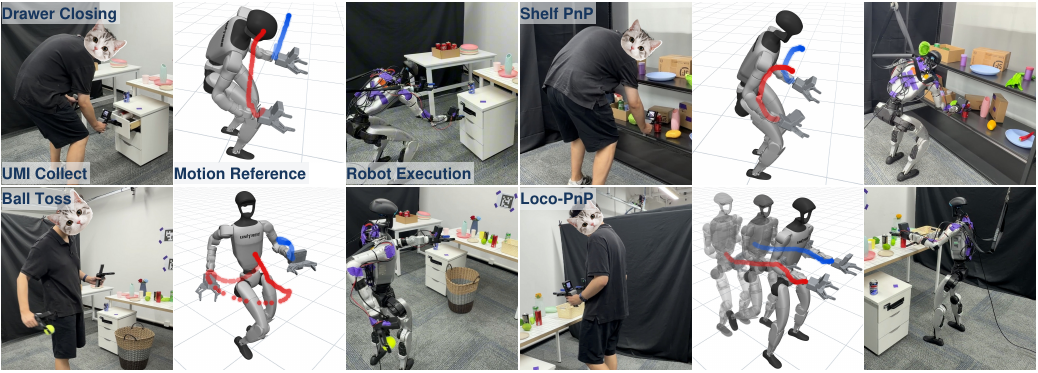}\\[\abovecaptionskip]
\refstepcounter{figure}\label{fig:teaser}%
\parbox{1\textwidth}{\normalfont\footnotesize\noindent
Fig.~\thefigure. \textbf{Whole-Body UMI: Universal Manipulation Interface with a Whole-Body Motion Prior}.
We show the UMI manipulation skill transfer process for four real-robot tasks:
drawer closing, shelf pick-and-place (PnP), ball toss, and locomotion pick-and-place (Loco-PnP). For each task, the panels
from left to right show UMI end-effector data collection, generated whole-body references,
and real-robot execution.
The hierarchy enables a real humanoid to perform whole-body manipulation,
including dynamic and locomotion-based tasks, using skills learned from native
UMI demonstrations.}%
\end{minipage}%
}

\begin{document}

\maketitle
\pagestyle{empty}

\begin{abstract}
Collecting whole-body demonstrations for humanoid manipulation mostly relies on teleoperation, which
is costly and hard to scale up. The Universal Manipulation Interface (UMI) provides a scalable data collection paradigm,
but end-effector trajectories alone underdetermine humanoid whole-body coordination, which is insufficient for whole-body demonstration collection.
Therefore, we introduce \textbf{Whole-Body UMI (WB-UMI)}, a task-agnostic, real-time and end-effector conditioned motion generator that
decouples whole-body coordination learning from task semantics
 learning through a shared end-effector
interface. A diffusion policy learns from native UMI demonstrations, while
\textbf{WB-UMI} learns independently from retargeted motion capture, requiring
no body trackers or paired image--whole-body demonstrations during task-specific data
collection. In real deployment, an asynchronous
hierarchy integrates the diffusion policy, motion generator, and a whole-body controller
with latency compensation and measured-state feedback. Real-robot experiments on G1 support real-time closed-loop
transfer across four tasks, achieving 90\% success in drawer closing, 80\% in
shelf pick-and-place, 30\% in ball toss, and 40\% in Loco-PnP, which shows the effectiveness of this hierarchy in
transferring native UMI skills to humanoid whole-body manipulation.
\end{abstract}

\section{Introduction}
Humanoid whole-body manipulation is valuable for many tasks in human
environments. Achieving precise and natural-like humanoid whole-body
manipulation requires
whole-body coordination data, consisting of task-centric inputs
(e.g., wrist-camera images to end-effector trajectories)
and complex action spaces (e.g., movements of the arms, torso, and legs).
One of the most effective ways is to learn from human
demonstrations.
However, most existing human data collection paradigms use teleoperation to
obtain whole-body
coordination through task-specific whole-body demonstrations, which is
laborious, costly, and hard to scale up.
The Universal Manipulation Interface (UMI) enables scalable, robot-free
collection of task-centric end-effector (EE) demonstrations~\cite{umi},
offering a promising way to reduce this data collection burden.

However, EE trajectories specify task intent but fundamentally underconstrain
humanoid whole-body coordination: the same EE trajectories can be realized through
different body postures and foot placements. How to generate plausible and effective
whole-body coordination from UMI data therefore still remains an open
challenge.
Existing solutions add body trackers to UMI and then use inverse kinematics
(IK) with hand-designed constraints to obtain coordinated whole-body
motion~\cite{humi,bifrostumi}. This process provides direct coordination
supervision, but introduces two limitations: 
\textit{(1) Additional instrumentation:} Users must wear and calibrate
multiple trackers in addition to the handheld UMI device, compromising the
convenience of UMI. \textit{(2) Limited scalability:} Although UMI with trackers improves
collection efficiency over teleoperation, it still requires task-specific
whole-body demonstrations. This dependence remains a fundamental bottleneck to
scaling data collection for rich and diverse whole-body coordination.

\textit{Can we leverage UMI's robot-free collection paradigm to avoid costly whole-body
demonstrations and enable scalable humanoid whole-body manipulation at the same time?}
Our key idea is to decouple task semantics learning from whole-body coordination
learning through a shared EE trajectory interface.
% Existing mocap data
% inherently encode correlations between EE poses and whole-body
% motions. We therefore learn task semantics from UMI demonstrations and a
% whole-body motion prior independently from mocap.
Therefore, we introduce \textbf{Whole-Body UMI (WB-UMI)}, a task-agnostic real-time and  EE-conditioned motion generator
trained on mocap independently. In our hierarchy (Fig.~\ref{fig:pipeline}(a)), a diffusion
policy (DP) trained on native UMI demonstrations predicts EE trajectories.
\textbf{WB-UMI} combines these trajectories with fused whole-body history
to generate explicit whole-body references for a generalist whole-body controller
(WBC). This proposed hierarchy connects task semantics learning to
whole-body execution without
additional task-specific paired whole-body motion collection and is validated across four tasks, achieving real-time, closed-loop execution
(Fig.~\ref{fig:teaser}).
In summary, our contributions are as follows:

\textit{(1) A new framework that decouples task semantics learning from whole-body
coordination learning:}
We decouple scalable UMI data collection from costly humanoid demonstrations by
learning whole-body coordination independently from existing mocap and connecting
it to task-specific skill learning through a shared end-effector (EE) interface.
This preserves UMI's convenience
while supporting scalable learning of rich humanoid whole-body coordination.

\textit{(2) Novel design of WB-UMI, a task-agnostic, real-time and EE-conditioned
whole-body motion generator for humanoids:}
\textbf{WB-UMI} enables whole-body motion generation from EE-only targets,
without additional body-keypoint constraints. It learns a rich, task-agnostic
whole-body motion prior from existing mocap that generalizes to unseen action categories
(Table~\ref{tab:offline}). It further combines low-latency streaming generation
with measured-state feedback from the robot for real-time deployment.

\textit{(3) A novel three-layer system for asynchronous closed-loop deployment:}
We integrate the task-specific DP, \textbf{WB-UMI}, and WBC for real-time
execution from images to whole-body actions. The system compensates for
generation latency, coordinates replanning through controller feedback, and
preserves manipulation targets during robot motion
(Fig.~\ref{fig:deploy-timeline}). In simulation, measured-state feedback
improves EE orientation and whole-body reference tracking
(Table~\ref{tab:tracker-execution}).
On the real robot, the integrated system achieves success rates of
90\%, 80\%, 30\%, and 40\%
on drawer closing, shelf PnP, ball toss, and Loco-PnP, respectively
(Fig.~\ref{fig:robot-demo-overview}).

\section{Related Work}

In this section, we review UMI data collection, conditioned motion generation, and Woble-body Control (WBC).

\subsection{UMI and the Humanoid Whole-Body Gap}
\label{sec:UMI-Gap}

Demonstration collection for humanoid manipulation differs in how it acquires
task semantics and whole-body coordination. Teleoperation records paired image--whole-body coordination under human
guidance~\cite{ben2025homie,ze2025twist,twist2}, yielding robot-native
demonstrations. However, collecting such data
is laborious, costly, and hard to scale up to diverse environments.
UMI~\cite{umi,fastumi}, instead, collects wrist-mounted images and EE trajectories directly from human demonstrations, offering a promising
approach to addressing scalability challenges. This robot-free collection paradigm has been
extended to quadruped manipulator
controllers~\cite{umionlegs}, as well as to bimanual mobile manipulator~\cite{hommi}, which proves the effectiveness of the UMI interface.
However, for high-DoF floating-base humanoid robots~\cite{darvish2023teleoperation},
EE trajectories alone leave whole-body motion underconstrained.
Recent robot-free interfaces address this gap by recording pelvis and foot
motion alongside EE trajectories, then retargeting these sparse keypoints through
inverse kinematics (IK)~\cite{humi,bifrostumi}. These interfaces retain robot-free
collection while supplying body targets for retargeting and control. However,
their coordination data must still be collected alongside task demonstrations.
In contrast, WB-UMI learns a reusable whole-body motion prior from existing
mocap, thereby decoupling whole-body coordination learning from
task-specific UMI demonstrations while preserving UMI's native EE-only
collection interface.

\subsection{Conditioned Motion Generation}
\label{sec:motion_generation}

Generative models learn reusable whole-body motion priors from mocap, supporting
diverse synthesis and adaptation to new
conditions~\cite{mdm,humanmotiondiffusion}. Spatial control builds on these priors
to make generated motion follow specified targets. Guidance during diffusion
sampling enforces global trajectories or sparse joint-position
constraints~\cite{gmd,omnicontrol}. Controllability has also been extended to
masked motion models with sparse and dense joint-position
constraints~\cite{maskcontrol}. Directly conditioned models incorporate position
and orientation targets into generation, supporting flexible keyframes and
trajectories for motion authoring~\cite{kimodo}.

Temporal continuity poses a complementary challenge for online use.
Autoregressive generators synthesize successive short motion segments from
history to accommodate changing commands~\cite{dartControl}. Larger-context
models additionally condition on future goals beyond the current prediction
window~\cite{zhao2026ardy}. Recent humanoid systems combine language-conditioned
streaming generation with low-level tracking for real-robot
execution~\cite{xie2026textoprealtimeinteractivetextdriven}. For UMI transfer, the
generator must follow predicted EE position and orientation trajectories while
adapting to execution errors. \textbf{WB-UMI} combines native EE-only conditioning,
sparse look-ahead context, and measured-state feedback to generate explicit
whole-body references online, without additional future body-keypoint targets
(Table~\ref{tab:capability}).

\begin{table}[!t]
\centering
\caption{Motion generator capabilities. EE-only: separate world-space
position/rotation constraints without extra future body targets. Real-time:
reported interactive generation. Stream.: continuous generation of motion
segments. FB: measured-state feedback. \textsuperscript{*}Spatial control through
inference-time optimization.}
\label{tab:capability}
\setlength{\tabcolsep}{0.5pt}
\renewcommand{\arraystretch}{1.10}
\begin{tabular*}{\columnwidth}{@{\extracolsep{\fill}}l*{8}{c}@{}}
\toprule
& \multicolumn{2}{c}{Online}
& \multicolumn{2}{c}{EE-only}
& \multicolumn{2}{c}{Robot}
& \multicolumn{2}{c}{Context (s)} \\
\cmidrule(lr){2-3}\cmidrule(lr){4-5}\cmidrule(lr){6-7}
\cmidrule(l){8-9}
Method & Real-time & Stream. & Pos. & Rot.
       & FB & \shortstack{Real\\robot}
       & Hist. & Future \\
\midrule
MaskControl (full)~\cite{maskcontrol}
    & \no  & \no  & \yes\textsuperscript{*} & \no  & \no  & \no  & \nap & 10.00 \\
Kimodo~\cite{kimodo}
    & \no  & \no  & \no & \yes & \no  & \no  & \nap & 10.00 \\
DartControl~\cite{dartControl}
    & \yes & \yes & \yes\textsuperscript{*} & \no  & \no  & \no  & 0.07 & 0.27 \\
ARDY~\cite{zhao2026ardy}
    & \yes & \yes & \no & \yes & \no  & \no  & 8.00 & 10.00 \\
\textbf{WB-UMI (ours)}
    & \yes & \yes & \yes & \yes & \yes & \yes & 0.27 & 2.13 \\
\bottomrule
\end{tabular*}
\end{table}

% Queue the two-column overview early so it appears on the Method opening page.
\begin{figure*}[!t]
\centering
\includegraphics[width=\textwidth]{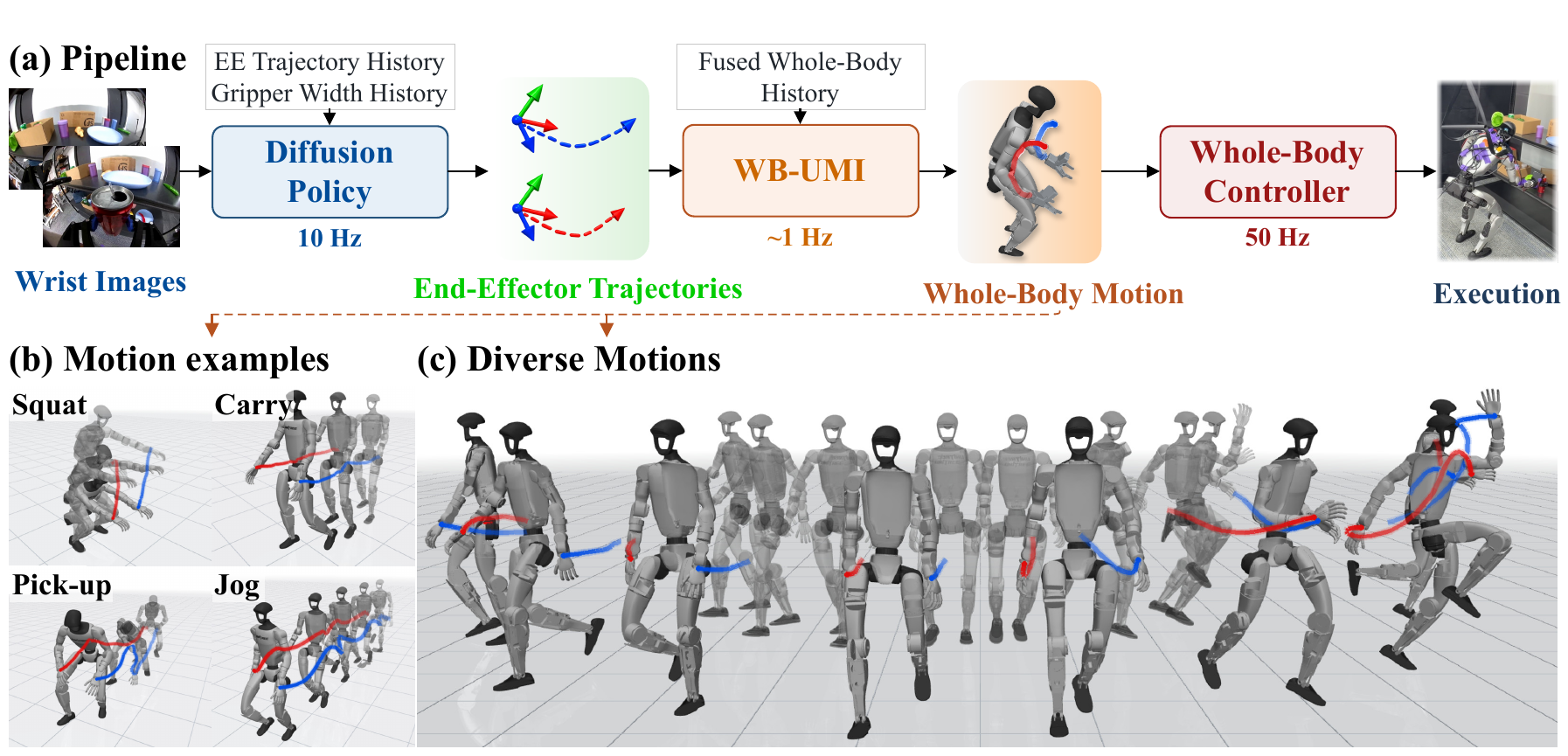}
\caption{Closed-loop deployment and generated motions. \textbf{(a)} A 10-Hz DP
supplies EE trajectories to \textbf{WB-UMI} ($\sim$1~Hz), whose references are
tracked by a 50-Hz WBC. \textbf{(b)} Squat, carry, pick-up, and jog motions.
\textbf{(c)} Motions generated from EE trajectories at different speeds.
Red/blue curves mark EE trajectories. The EE interface supports diverse whole-body coordination within a
closed-loop visual control hierarchy. 
% [motion synthesis]
}
\label{fig:pipeline}
\end{figure*}

\subsection{Whole-Body Control}
\label{sec:wbc}
Classical whole-body control (WBC) often relies on hierarchical quadratic
programming (QP) to coordinate multiple tasks under kinematic and dynamic
constraints~\cite{escande2014hierarchical,herzog2016momentum}.
Reinforcement learning (RL)-based motion trackers have been developed for
simulated characters and humanoid
robots~\cite{peng2018deepmimic,zhang2025trackmotionsdisturbances,beyondmimic}.
Scaling motion data and policy capacity further enables generalist WBC
that tracks unseen motions and supports multiple command interfaces
with a single policy~\cite{sonic}.
Alongside generalist tracking, behavior foundation models broaden task
specification through shared latent representations of motions, goals, and
rewards~\cite{li2025bfmzeropromptablebehavioralfoundation} and unified
motion-command interfaces~\cite{zeng2026scalingbehaviorfoundationmodel}.
In contrast, \textbf{WB-UMI} converts EE-only UMI commands into explicit
whole-body references for a pretrained generalist WBC.

\section{Method}
\label{sec:method}
In this section, we introduce our method in detail, including the problem formulation, \textbf{WB-UMI}, and description of its integration into a closed-loop deployment system.

\subsection{Problem Formulation and System Overview}
We assume access to native UMI task demonstrations, an independent dataset
of retargeted mocap, and a whole-body controller, but no task-specific whole-body
demonstrations. Given observation history $\mathbf{o}_t$, comprising wrist
RGB images and proprioceptive measurements, we seek a closed-loop system that generates coordinated
whole-body actions in real time, enabling the humanoid to achieve the task-specific whole-body manipulation with only end-effector-level task demonstrations, where no future
other body-keypoint targets are provided.
Figure~\ref{fig:pipeline}(a) illustrates our three-layer system. A DP
trained on native UMI demonstrations maps $\mathbf{o}_t$ to EE trajectories and gripper widths. \textbf{WB-UMI}, trained
independently on mocap, combines these EE trajectories with
fused whole-body history to generate explicit whole-body kinematic
references. The pretrained SONIC~\cite{sonic}, which serves as our WBC here, tracks these references
to execute the action. Controller feedback updates the observation and
whole-body histories for subsequent predictions.

\subsection{\textbf{WB-UMI}: Real-Time EE-Conditioned Generation}
\begin{figure*}[t]
\centering
\makebox[\textwidth][c]{%
  \includegraphics[width=\textwidth]{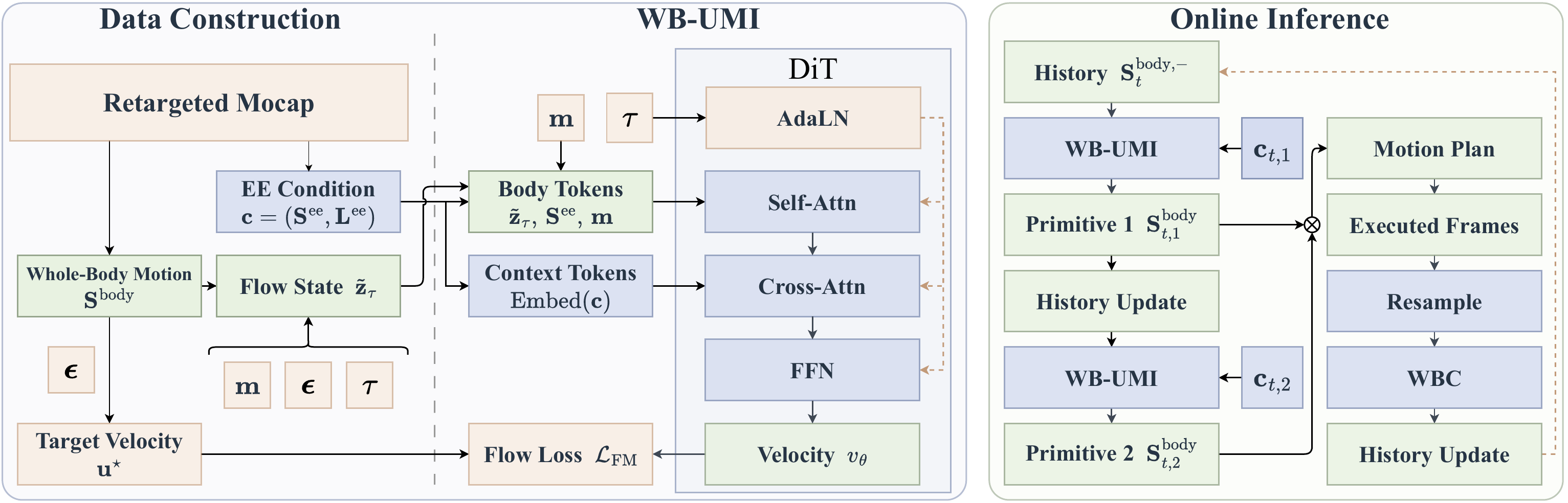}%
}
\caption{\textbf{WB-UMI learns an EE-conditioned whole-body motion prior from retargeted mocap and deploys it through closed-loop autoregressive generation.} Left: Retargeted mocap provides paired whole-body motions and EE conditions for training. Center: The \textbf{WB-UMI} architecture is trained with masked flow matching to predict an EE-conditioned velocity field. Right: Detailed inference process for a single motion segment with execution history feedback.}
\label{fig:motion-prior-architecture}
\end{figure*}

Given whole-body history and EE future trajectories, \textbf{WB-UMI}
infers coordinated whole-body motion without additional future body-keypoint
targets. We autoregressively generate long-horizon motion as a sequence of short, fixed-length primitives~\cite{xie2026textoprealtimeinteractivetextdriven}, enabling continuous real-time updates in response to evolving EE targets. Each primitive contains $F$
future frames conditioned on $H$ history frames:
\begin{equation}
\mathbf{S}^{{\rm body},+}
=G_\theta\!\left(\mathbf{S}^{{\rm body},-},
\mathbf{S}^{\rm ee},\mathbf{L}^{\rm ee}\right),
\label{eq:prior-mapping}
\end{equation}
where $\mathbf{S}^{\rm ee}$ supplies EE conditions aligned with
$\mathbf{S}^{\rm body}=[\mathbf{S}^{{\rm body},-},\mathbf{S}^{{\rm body},+}]$,
and $\mathbf{L}^{\rm ee}$ provides sparse look-ahead context. We write
$\mathbf{c}=(\mathbf{S}^{\rm ee},\mathbf{L}^{\rm ee})$. Two primitives
form a $2F$-frame motion segment. Table~\ref{tab:sequence-notation} defines the
notation, and Fig.~\ref{fig:motion-prior-architecture} summarizes the model.

\begin{table}[!htbp]
\centering
\caption{Core notation for \textbf{WB-UMI}.}
\label{tab:sequence-notation}
\setlength{\tabcolsep}{3pt}
\renewcommand{\arraystretch}{0.92}
\begin{tabularx}{\columnwidth}{@{}L{0.22\columnwidth}Xr@{}}
\toprule
\textbf{Notation} & \textbf{Description} & \textbf{Value} \\
\midrule
$-$ / $+$ & History / future segment & -- \\
$i,t,k$ & Frame (30~Hz), replanning step, and primitive
index ($k\in\{1,2\}$) & -- \\
$H$ / $F$ & History / prediction length (frames) & 8 / 32 \\
$K$ / $d$ & Look-ahead length / sampling stride (frames) & 32 / 4 \\
$\operatorname{Sample}_d$ & Stride-$d$ sampling from the first frame;
8 EE samples per look-ahead window & -- \\
$\mathbf{c}_{t,k}$ & EE condition for primitive $k$ at step $t$ & -- \\
$\mathbf{S}_{t,k}^{\rm body}$ & Generated $F$-frame primitive $k$ at step $t$ & -- \\
\bottomrule
\end{tabularx}
\end{table}

\par\noindent\textbf{Data Construction.}
For training, we construct paired whole-body states $\mathbf{s}_i^{\rm body}$ and EE
conditions $\mathbf{s}_i^{\rm ee}$ entirely from retargeted G1 mocap~\cite{bonesseed},
extracting EE trajectories through forward kinematics (FK).
Table~\ref{tab:method-representation} summarizes both representations.

The whole-body state $\mathbf{s}_i^{\rm body}$ builds on a local incremental robot
representation~\cite{xie2026textoprealtimeinteractivetextdriven}, making
frame-to-frame changes explicit while retaining absolute root height.
The EE condition $\mathbf{s}_i^{\rm ee}$ combines relative poses, pose increments, and initial height and orientation anchors.
Both representations are anchored at the first history frame of each segment. Root planar motion and EE displacements are measured relative to their respective initial positions and rotated by the inverse initial
root yaw. EE rotations are relative to their initial orientations, and orientation anchors are expressed in the initial root heading frame.
This local formulation reduces dependence on global placement.

\begin{table}[!htbp]
\centering
\caption{Representations of the whole-body state and EE condition.}
\label{tab:method-representation}
\setlength{\tabcolsep}{3pt}
\renewcommand{\arraystretch}{0.92}
\begin{tabularx}{\columnwidth}{@{}lXr@{}}
\toprule
Component & Representation & Size \\
\midrule
\multicolumn{2}{@{}l}{\textit{Whole-body state $\mathbf{s}^{\rm body}_i$}} &
\textbf{Sum: 73} \\
Root orientation & Roll/pitch sine--cosine; yaw increment & 5 \\
Root position & Translation increment, height, planar pose & 8 \\
Joint motion & Joint positions and increments & 58 \\
Foot contact & L/R: ankle $h<0.08$\,m, $v<0.1$\,m/s & 2 \\
\midrule
\multicolumn{2}{@{}l}{\textit{EE condition $\mathbf{s}_i^{\rm ee}$}} &
\textbf{Sum: 50} \\
Relative pose & Segment-relative position and 6D orientation & 18 \\
Height anchor & Initial heights & 2 \\
Orientation anchor & Initial heading-frame 6D orientations & 12 \\
Pose increment & Position and orientation increments & 18 \\
\bottomrule
\end{tabularx}
\end{table}

At inference, the DP provides EE trajectories extending beyond the current primitive. The segment aligned with the generated frames serves as tracking targets, while less reliable farther predictions are sparsely sampled as look-ahead context to convey motion intent. The resulting EE condition is:
\begin{equation}
\begin{aligned}
\mathbf{S}^{\rm ee}
  &=(\mathbf{s}^{\rm ee}_i)_{i=1}^{H+F},\\
\mathbf{L}^{\rm ee}
  &=\operatorname{Sample}_d\!\left(
    (\mathbf{s}^{\rm ee}_i)_{i=H+F+1}^{H+F+K}\right).
\end{aligned}
\label{eq:ee-condition}
\end{equation}

\par\noindent\textbf{Training.}
We train \textbf{WB-UMI} on pairs of consecutive primitives using masked
conditional flow matching. During training, we gradually replace the second
primitive's ground-truth history with generated motion from the first.
For each primitive, we construct the input at flow time $\tau\in[0,1]$ using
Gaussian noise $\boldsymbol\epsilon$:
\begin{equation}
\tilde{\mathbf z}_\tau
=\big[\mathbf S^{{\rm body},-},\,(1-\tau)\boldsymbol\epsilon
+\tau\mathbf S^{{\rm body},+}\big],
\label{eq:masked-flow-path}
\end{equation}
where the history remains fixed, while the future segment transitions linearly
from noise at $\tau=0$ to ground-truth motion at $\tau=1$.
The binary mask $\mathbf m$ marks history frames with $1$ and future frames
with $0$.
The conditional DiT $v_\theta$ predicts flow velocity using two EE
conditioning paths (Fig.~\ref{fig:motion-prior-architecture}). It forms each
body token from a frame of $\tilde{\mathbf z}_\tau$ and the time-aligned EE
features in $\mathbf S^{\rm ee}$. Self-attention models temporal dependencies
among these tokens, while cross-attention provides access to the full EE
context $\mathbf c$. For auxiliary supervision, we use the predicted velocity
to estimate the clean future:
\begin{equation}
\widehat{\mathbf S}^{{\rm body},+}
=\tilde{\mathbf z}_\tau^+
+(1-\tau)v_\theta^+(\tilde{\mathbf z}_\tau,\tau,\mathbf c).
\label{eq:clean-estimate}
\end{equation}

The flow-matching loss supervises each primitive's future velocity against
the target flow velocity $\mathbf u^\star=\mathbf S^{{\rm body},+}-\boldsymbol\epsilon$:
\begin{equation}
\mathcal L_{\rm FM}=\mathbb E\!\left\|
v_\theta^+(\tilde{\mathbf z}_\tau,\tau,\mathbf c)
-\mathbf u^\star
\right\|_2^2,
\label{eq:masked-flow-loss}
\end{equation}
where the expectation is over training windows, flow times, and noise.
For all auxiliary losses below, $\mathbf S^{{\rm body},+}$ and
$\widehat{\mathbf S}^{{\rm body},+}$ denote the concatenated ground-truth
futures and clean estimates of both primitives. EE targets are frame-aligned,
and geometric comparisons use a common coordinate frame.
The state loss combines reconstruction and root consistency:
\begin{equation}
\mathcal L_{\rm state}
=\|\widehat{\mathbf S}^{{\rm body},+}-\mathbf S^{{\rm body},+}\|_2^2
+\ell_{\rm root},
\label{eq:state-loss}
\end{equation}
where $\ell_{\rm root}$ encourages accurate root placement and internally
consistent root motion.
The EE loss compares FK-derived relative poses with the frame-aligned targets:
\begin{equation}
\mathcal{L}_{\rm EE}=\|\widehat{\mathbf S}^{\rm ee}_{\rm pose}-\mathbf S^{\rm ee}_{\rm pose}\|_2^2+\ell_{\rm rot}(\widehat R,R_c),
\label{eq:ee-tracking-loss}
\end{equation}
where $\ell_{\rm rot}$ penalizes orientation errors between predictions
$\widehat R$ and targets $R_c$.
We supervise body poses, velocities, and accelerations with
\begin{equation}
\mathcal{L}_{\rm coord}=\sum_{i=0}^{2}\|\mathcal D_i(\widehat{\mathbf S}^{{\rm body},+})-\mathcal D_i(\mathbf S^{{\rm body},+})\|_2^2,
\label{eq:coordination-loss}
\end{equation}
where $\mathcal D_0=\mathrm{FK}$ gives joint poses, including sole positions
at contact, while $\mathcal D_1=\Delta$ and $\mathcal D_2=\Delta^2$ compute
first and second frame differences of body features. Temporal terms emphasize
root motion and primitive boundaries. Foot support is supervised by
\begin{equation}
\mathcal{L}_{\rm skate}=\|\widehat{\mathbf v}_f\|_{\mathcal C}^2-\log p_c,
\label{eq:skate-loss}
\end{equation}
where $\widehat{\mathbf v}_f$ is predicted sole velocity, $\mathcal C$ weights
recorded or predicted contact, and $p_c$ is the predicted probability of the
recorded foot-contact label.
The training objective is
\begin{equation}
\mathcal{L}=\mathcal{L}_{\rm FM}+\mathcal{L}_{\rm state}+\mathcal{L}_{\rm EE}+\mathcal{L}_{\rm coord}+\mathcal{L}_{\rm skate}.
\label{eq:total-method-loss}
\end{equation}
\par\noindent\textbf{Inference.}
For each primitive, we keep $\mathbf S^{{\rm body},-}$ and $\mathbf c$ fixed,
initialize the future segment with $\boldsymbol\epsilon$, and integrate
the flow using $v_\theta^+$ from $\tau=0$ to $1$ in eight uniform Euler steps.
To generate $\mathbf S_{t,2}^{\rm body}$, we use the last $H$ frames of
$\mathbf S_{t,1}^{\rm body}$ as history and advance the EE window by $F$
frames to obtain $\mathbf c_{t,2}$. We then repeat the same sampling procedure
in the shared local coordinate frame.
\label{sec:deployment}
\begin{figure}[!t]
\centering
\includegraphics[width=\columnwidth]{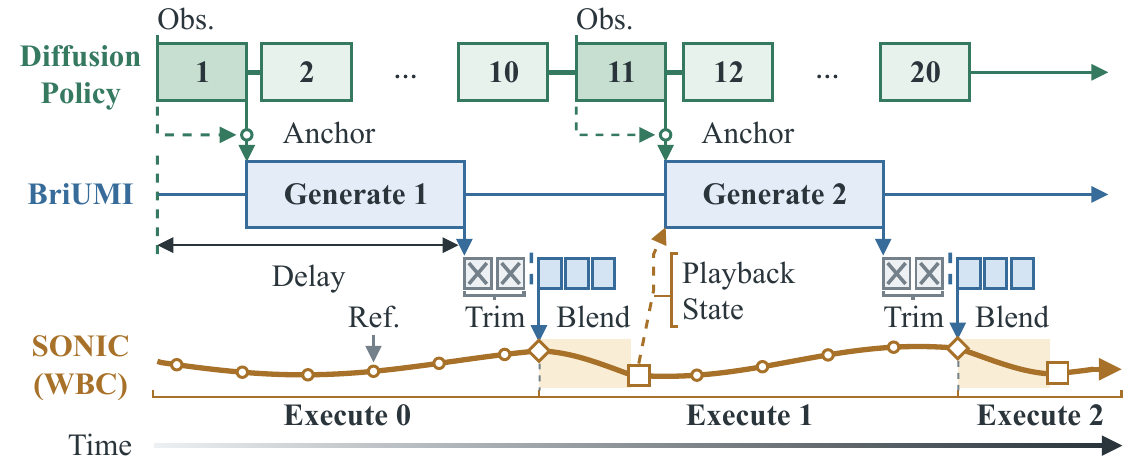}
\caption{Asynchronous deployment. \textbf{WB-UMI} generates new motion
while the WBC tracks the active reference. Incoming references are
trimmed and blended based on playback progress.}
\label{fig:deploy-timeline}
\end{figure}
\subsection{Integrated Deployment System}
Figure~\ref{fig:deploy-timeline} shows the asynchronous planning and execution
loop. To keep EE targets fixed as the robot moves, we anchor each relative DP
prediction to the hand pose at its source observation:
\begin{equation}
\mathbf T_{\rm target}=\mathbf T_{\rm obs}\mathbf T_{\rm rel},
\label{eq:deployment-anchor}
\end{equation}
where $\mathbf T_{\rm obs}$ is the hand pose at that observation and
$\mathbf T_{\rm rel}$ is the predicted relative EE pose.
\textbf{WB-UMI} uses these anchored targets and the current whole-body history to
generate a new motion reference. Once it is ready, we trim the prefix that
can no longer be executed and blend the remaining reference into ongoing
execution:
\begin{equation}
\mathbf q_{\rm cmd}[j]=\operatorname{Blend}_{w_j}\!\left(\mathbf q_{\rm old}[j],\mathbf q_{\rm new}[j+\delta]\right),
\label{eq:deployment-trim-blend}
\end{equation}
where $\delta$ is the frame offset aligning the incoming reference
$\mathbf q_{\rm new}$ with the active reference $\mathbf q_{\rm old}$
in execution time, and the blend weight $w_j$ gradually increases
from zero to one.
Replanning advances by $F$ planner frames, with the second primitive serving
as a preview. At each planning boundary, controller feedback confirms the
active reference and playback progress. We then update the whole-body history for
the next \textbf{WB-UMI} generation:
\begin{equation}
\mathbf S_{t+1}^{{\rm body},-}=\mathbf m_{\rm fb}\odot\mathbf S_{\rm meas}^{-}+(1-\mathbf m_{\rm fb})\odot\mathbf S_{\rm ref}^{-},
\label{eq:deployment-feedback}
\end{equation}
where $\mathbf S_{\rm meas}^{-}$ and $\mathbf S_{\rm ref}^{-}$ are measured
and reference histories aligned in time and coordinates. The feedback mask
$\mathbf m_{\rm fb}$ selects measured channels, including joints and base tilt,
and retains reference values elsewhere.

\newcounter{expheadnumber}[subsection]
\newcommand{\exphead}[1]{\par\noindent\hspace*{\parindent}\stepcounter{expheadnumber}\textit{(\arabic{expheadnumber}) #1:}\ }

\section{Experiments}
\label{sec:exp}

In this section, we evaluate \textbf{WB-UMI} and the proposed hierarchy through the following questions:
\begin{enumerate}
\item[\textbf{Q1:}] \textbf{Motion generation quality.} How well does \textbf{WB-UMI}
generate coordinated whole-body motion from EE trajectories on unseen action
categories?
\item[\textbf{Q2:}] \textbf{Execution performance.} How accurately are the
generated references tracked in simulation?
\item[\textbf{Q3:}] \textbf{UMI skill transfer.} Can the proposed hierarchy
enable whole-body manipulation on a physical humanoid from native UMI
demonstrations, without task-specific whole-body demonstrations?
\end{enumerate}

\subsection{Motion Generation Quality}
\label{sec:exp-offline}

\exphead{Setup}
Using the settings in Table~\ref{tab:model-configurations}, we evaluate
\textbf{WB-UMI} offline on EE trajectories extracted from mocap data, with
generated motion providing the whole-body history during autoregressive rollout.
The G1 dataset is built from BONES-SEED~\cite{bonesseed}, retaining basic
locomotion and manipulation while excluding complex motions that rarely
occur in whole-body manipulation. It contains approximately 105 hours:
90 for training, 5 for validation, and 10 for testing. We split clips by
semantic action group (e.g., “axe splitting wood”), with no group
shared across splits, so the test categories are unseen during training.
% Queue the two-column demo figure early so it appears at the top of page 6.
\newsavebox{\robotdemorowbox}
\newlength{\robotdemorowheight}
\begin{figure*}[!t]
\centering
\sbox{\robotdemorowbox}{\includegraphics[width=0.72\textwidth]{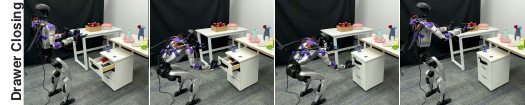}}%
\setlength{\robotdemorowheight}{\dimexpr\ht\robotdemorowbox+\dp\robotdemorowbox\relax}%
\makebox[\textwidth][c]{%
\begin{minipage}[t]{1.0\textwidth}
\begin{minipage}[t]{0.72\linewidth}
  \vspace{0pt}
  \centering
  {\normalfont\sffamily\normalsize\bfseries Real-Robot
  Execution\par}
  \vspace{2pt}

  \includegraphics[width=\linewidth]{figures/experiment_demo_drawer_contrast_q91.pdf}%
  \par\nointerlineskip\vspace{0.8pt}%
  \includegraphics[width=\linewidth]{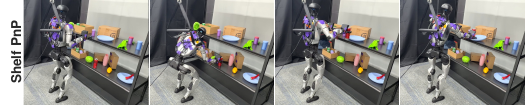}%
  \par\nointerlineskip\vspace{0.8pt}%
  \includegraphics[width=\linewidth]{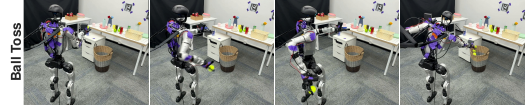}%
  \par\nointerlineskip\vspace{0.8pt}%
  \includegraphics[width=\linewidth]{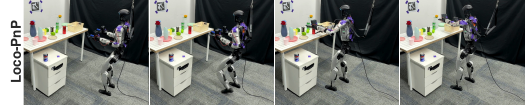}%
  \par\nointerlineskip\vspace{1pt}%
  \makebox[\linewidth][l]{%
    \makebox[0.045174\linewidth][c]{$t$}%
    \raisebox{0.25ex}{\includegraphics[width=0.954826\linewidth]{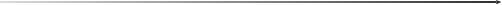}}%
  }
\end{minipage}\hspace{2pt}%
\begin{minipage}[t]{0.265\linewidth}
  \vspace{0pt}
  \raggedright
  \makebox[0.93\linewidth][c]{\normalfont\sffamily\normalsize\bfseries
  Quality Analysis}\par
  \vspace{0pt}

  \includegraphics[height=\robotdemorowheight]{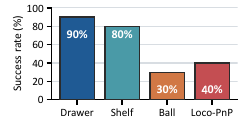}%
  \par\nointerlineskip\vspace{0.8pt}%
  \includegraphics[height=\robotdemorowheight]{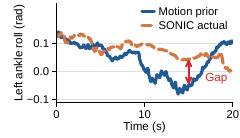}%
  \par\nointerlineskip\vspace{0.8pt}%
  \includegraphics[height=\robotdemorowheight,trim=0 114 0 8,clip]{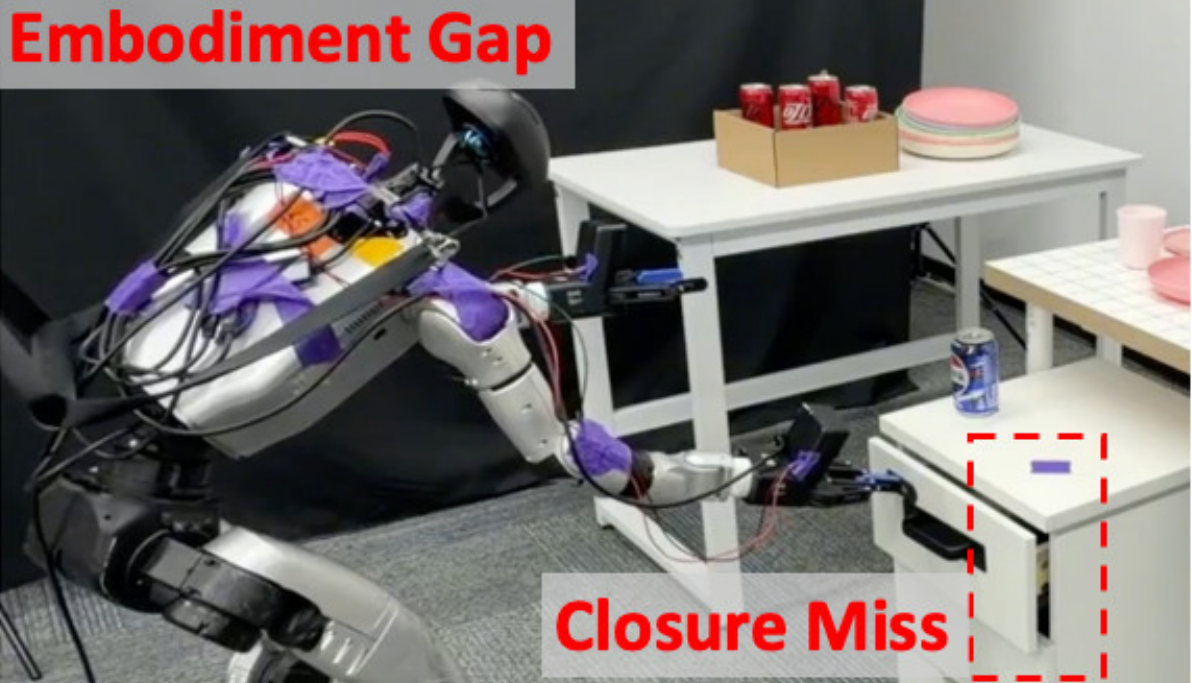}%
  \par\nointerlineskip\vspace{0.8pt}%
  \includegraphics[height=\robotdemorowheight]{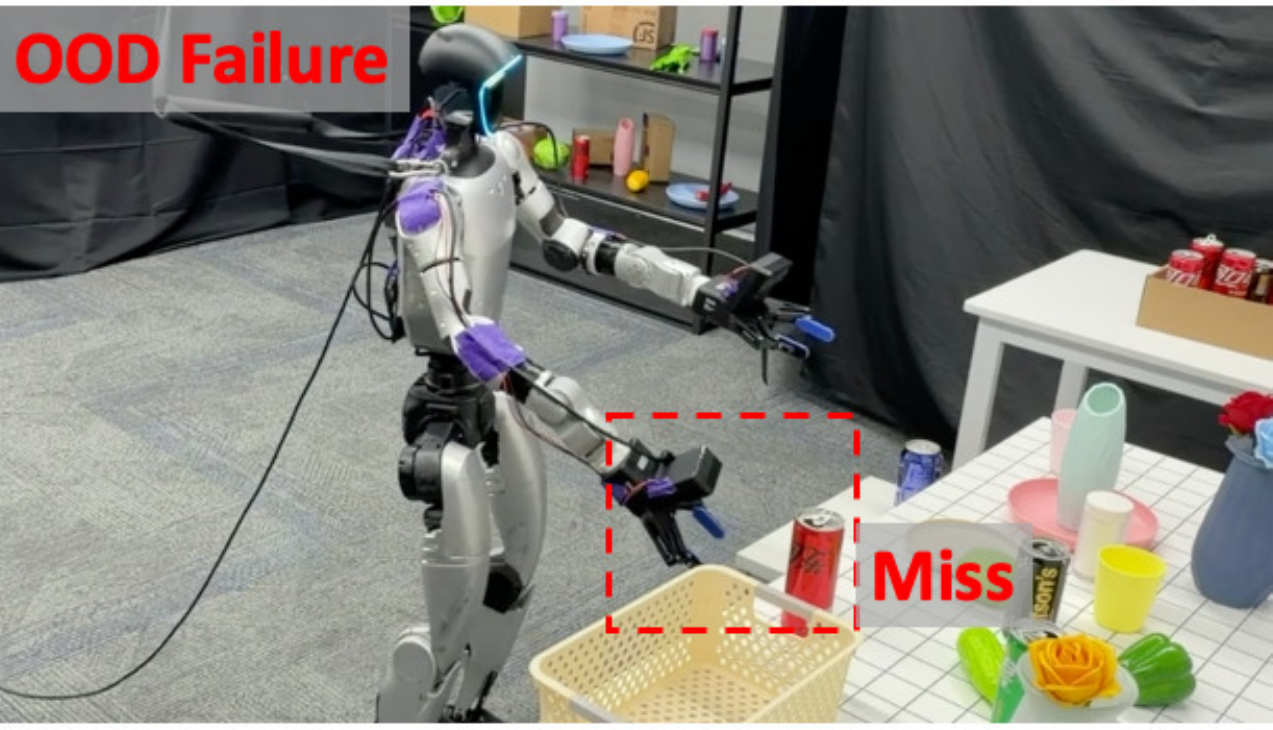}
\end{minipage}
\end{minipage}%
}
\caption{Real-robot transfer with \textbf{WB-UMI}. Left: drawer closing, shelf
PnP, ball toss, and Loco-PnP. Right (top to bottom): success rates, left ankle
roll tracking, and drawer-closing and Loco-PnP failures. Solid blue and dashed
orange curves show the \textbf{WB-UMI} reference and SONIC execution,
respectively. Red annotations mark tracking errors and failures.
These results demonstrate successful UMI skill transfer, with remaining
limitations in reachability and whole-body tracking.}
\label{fig:robot-demo-overview}
\end{figure*}

\begin{table}[!t]
\centering
\caption{Configurations of the DP and \textbf{WB-UMI}.}
\label{tab:model-configurations}
\setlength{\tabcolsep}{3pt}
\renewcommand{\arraystretch}{0.92}
\begin{tabularx}{\columnwidth}{@{}L{0.43\columnwidth}X@{}}
\toprule
\textbf{Setting} & \textbf{Value} \\
\midrule
\multicolumn{2}{@{}l}{\textbf{DP}} \\
Input horizons (vis. / prop.) & 2 / 3 \\
Observation / action frequency & 10 / 10~Hz \\
Action horizon & 40 \\
Visual input / encoder & $2\times224\times224$; DINOv3 ViT-B/16 \\
DP backbone & DiT; 6 blocks, width 384, 6 heads \\
Learning rates (DP / vision) & $10^{-4}$ / $10^{-5}$ \\
Training & 100 epochs; batch 16; flow matching \\
\midrule
\multicolumn{2}{@{}l}{\textbf{WB-UMI}} \\
History / prediction horizon & 8 / 32 \\
Rollout & 2 primitives at 30~Hz \\
EE look-ahead horizon / stride & 32 / 4 \\
Temporal backbone & DiT; 10 blocks, width 384, 6 heads \\
Learning rate & $5.66\times10^{-4}$ \\
Training & 300 epochs; batch 8192; masked flow matching \\
Inference & 8 ODE steps \\
\bottomrule
\end{tabularx}
\end{table}

\exphead{Comparisons}
We compare \textbf{WB-UMI} with a variant without EE look-ahead and oracle
variants given ground-truth root poses or root and foot poses. All variants
are retrained using the same data and backbone and evaluated with the same
sampling procedure. The oracle conditions cover the same history and
prediction frames as the EE conditions, excluding look-ahead. These additional
body signals are unavailable in native UMI demonstrations.

\exphead{Metrics}
EE position, velocity, and orientation errors measure tracking of the input
trajectories. Joint-rotation and FK joint-position errors measure reconstruction
of the recorded whole-body motion, which is one possible realization of the EE
constraints. Foot velocity during contact measures sliding. These framewise
metrics are averaged within each clip and then across clips.
% \todo{Supplementary metric definitions: specify error norms, coordinate alignment, and whether sliding uses recorded or generated contact masks.}
FID measures distributional similarity between generated and recorded motions
in a learned feature space. The encoder is trained only on real motions from
the training split and kept fixed during evaluation. Using equally weighted
clip features, we compute
\begin{equation}
\mathrm{FID}=\|\boldsymbol{\mu}_g-\boldsymbol{\mu}_r\|_2^2+
\operatorname{Tr}\!\left(\boldsymbol{\Sigma}_g+\boldsymbol{\Sigma}_r-
2(\boldsymbol{\Sigma}_g\boldsymbol{\Sigma}_r)^{1/2}\right),
\label{eq:motion-fid}
\end{equation}
where $\boldsymbol{\mu}$ and $\boldsymbol{\Sigma}$ are the feature mean and
covariance, and $g$ and $r$ denote generated and recorded motions, respectively.
% FID implementation detail retained for supplementary reporting: a frozen
% temporal convolutional autoencoder encodes standardized, heading-aligned
% windows. Window features are L2-normalized, averaged within each clip, and
% normalized again. FID uses the means and covariances of the resulting clip features.
% \todo{Supplementary reproducibility: provide the FID encoder configuration, training objective, and window length/stride.}

\exphead{Results}
On unseen action categories, \textbf{WB-UMI} achieves an EE position error of
4.25~cm during autoregressive rollout (Table~\ref{tab:offline}).
Figure~\ref{fig:pipeline}(b,c) illustrates leg and trunk coordination across
manipulation and locomotion, as well as changes in whole-body motion with EE
trajectory speed. The oracle comparisons show different effects of additional
body constraints: root and foot conditions together improve body reconstruction
and reduce sliding relative to \textbf{WB-UMI}, whereas root conditions alone
increase sliding. Back to \textbf{Q1}:
\textbf{WB-UMI} generates coordinated whole-body motion from EE trajectories on
unseen action categories, while EE look-ahead improves all reported generation
metrics.
% Optional figure detail: identify held-out examples if using this figure as evidence of unseen-category generalization, and describe the EE speed variants in its caption or supplementary material.

\begin{table}[!h]
\centering
\caption{Offline generation on unseen action categories. Bold compares the
two EE-only variants; oracle rows assess additional body conditioning.}
\setlength{\tabcolsep}{0.5pt}
\renewcommand{\arraystretch}{1.10}
\begin{tabularx}{\columnwidth}{@{}
  >{\raggedright\arraybackslash}p{0.26\columnwidth}*{7}{C}@{}}
\toprule
Condition & \shortstack{EE\\pos.} & \shortstack{EE\\vel.}
       & \shortstack{EE\\orient.} & \shortstack{Foot\\vel.}
       & \shortstack{Joint\\rot.} & \shortstack{Joint\\pos.} & FID \\
       & (cm)$\downarrow$ & (m/s)$\downarrow$ & (deg)$\downarrow$
       & (m/s)$\downarrow$ & (deg)$\downarrow$ & (m)$\downarrow$
       & $\downarrow$ \\
\midrule
\textbf{WB-UMI} w/o look-ahead & 5.77 & 0.0438 & 2.52 & 0.0590 & 5.31
                         & 0.0583 & 0.0280 \\
\mbox{\textbf{WB-UMI}} & \textbf{4.25} & \textbf{0.0364} & \textbf{1.42}
          & \textbf{0.0536} & \textbf{4.27} & \textbf{0.0439}
          & \textbf{0.0202} \\
\midrule
EE+root (oracle) & 3.93 & 0.0439 & 2.43 & 0.0669
                 & 4.31 & 0.0289 & 0.0175 \\
EE+root+feet (oracle) & 3.87 & 0.0323 & 1.78 & 0.0236 & 2.56
                       & 0.0163 & 0.0079 \\
\midrule
Reference motion & - & - & - & 0.008 & - & - & - \\
\bottomrule
\end{tabularx}
\label{tab:offline}
\end{table}

\subsection{Simulated Execution Performance}
\label{sec:exp-execution}
\exphead{Setup}
We evaluate whole-body reference execution for shelf PnP in MuJoCo using
a single EE trajectory selected from UMI demonstrations collected with the
GenRobot DAS Gripper~\cite{genrobot_das_gripper}. Each method replays this
trajectory 100 times without the DP, using identical initial states and
safety limits and no external disturbances. All rollouts completed the
recorded trajectory without falls or early termination.

\exphead{Comparisons}
\textbf{WB-UMI} generates whole-body references for the pretrained SONIC
controller. Our EE-only baseline is \textit{Direct EE-WBC}, trained with
reinforcement learning using randomly sampled EE targets as task commands.
We also include IK (\texttt{Mink}~\cite{mink}) as a reference under a manually
specified, fixed support-contact configuration for shelf PnP. It supplies
whole-body references to the same SONIC controller, but requires additional
contact information beyond the EE commands, so it is not an EE-only baseline.
We compare \textbf{WB-UMI} w/o feedback and \textbf{WB-UMI} (full) using
the same trained checkpoint. The full variant updates its body history with
measured robot state at each replanning step, while the variant without
feedback uses body history from the previous motion prediction.
% \todo{Supplementary baseline configuration: document Direct EE-WBC training and the IK objectives, weights, and constraints.}

\exphead{Metrics}
EE velocity and orientation errors measure adherence to the shared commands.
Foot slip measures mean horizontal ankle-link speed during inferred stance.
Root-relative mean per-joint position error (MPJPE) and joint-rotation error
measure tracking of \emph{each method's own reference}. These two metrics
do not apply to Direct EE-WBC, which produces no whole-body reference.
We first average each metric over frames within each rollout, then compute
the mean and standard deviation across the 100 rollouts.
% Whole-body reference tracking metric details retained for supplementary reporting: MPJPE
% averages Euclidean joint-position errors after subtracting each motion's
% root position; joint-rotation error averages absolute angle differences
% over the 29 actuated joints.
% \todo{Supplementary metric definitions: specify EE error norms, command/reference time alignment, and stance detection, reusing Method definitions where applicable.}

\begin{table}[!ht]
\centering
\caption{Shelf PnP execution over 100 repetitions of a single UMI trajectory per
method (mean $\pm$ standard deviation; lowest means in bold).}
\setlength{\tabcolsep}{1.5pt}
\setlength{\medmuskip}{2mu}
\renewcommand{\arraystretch}{1.18}
\begin{tabular*}{\columnwidth}{@{\extracolsep{\fill}}lcccc@{}}
\toprule
Metric ($\downarrow$) & \shortstack{IK\\(ref.)} & \shortstack{Direct\\EE-WBC}
& \shortstack{\textbf{WB-UMI}\\w/o feedback}
& \shortstack{\textbf{WB-UMI}\\(full)} \\
\midrule
EE vel. (cm/s) & $\mathbf{5.40\pm0.50}$ & $17.49\pm6.50$ & $7.85\pm1.02$
& $8.36\pm1.11$ \\
EE orient. (deg) & $12.47\pm1.17$ & $38.89\pm16.41$ & $6.76\pm0.50$
& $\mathbf{4.97\pm0.47}$ \\
Foot slip (cm/s) & $2.30\pm1.30$ & $7.05\pm10.55$ & $\mathbf{0.87\pm0.54}$
& $0.96\pm0.63$ \\
\shortstack[l]{Root-rel.\\MPJPE (cm)} & $2.00\pm0.20$ & -- & $2.36\pm0.20$
& $\mathbf{1.73\pm0.16}$ \\
\shortstack[l]{Joint rot.\\err. (deg)} & $7.28\pm0.46$ & -- & $4.33\pm0.31$
& $\mathbf{2.65\pm0.16}$ \\
\bottomrule
\end{tabular*}
\label{tab:tracker-execution}
\end{table}

\exphead{Results}
On shelf PnP, both \textbf{WB-UMI} variants yield lower mean EE velocity and
orientation errors and less foot slip than Direct EE-WBC
(Table~\ref{tab:tracker-execution}). Measured-state feedback reduces whole-body
reference tracking and EE orientation errors, while slightly increasing EE
velocity error and foot slip. The IK reference yields the lowest EE velocity
error under a manually specified contact configuration, whereas \textbf{WB-UMI}
(full) yields lower EE orientation error and foot slip without this
specification. These results answer \textbf{Q2}: In the simulation, \textbf{WB-UMI} achieves lower EE velocity
and orientation errors and less foot slip than Direct EE-WBC. Measured-state
feedback improves whole-body reference tracking and EE orientation, with small
increases in EE velocity error and foot slip.

\subsection{UMI Skill Transfer}
We evaluate UMI skill transfer on four tasks with the 29-DoF G1 humanoid
(Fig.~\ref{fig:robot-demo-overview}), testing only the proposed hierarchy for
safety. For each task, we train the DP on 200 native UMI demonstrations
collected with the GenRobot DAS Gripper~\cite{genrobot_das_gripper} in
approximately 90 minutes. \textbf{WB-UMI} learns its whole-body motion prior
independently from mocap, requiring no task-specific whole-body demonstrations.
The G1 uses the corresponding DAS Controller~\cite{genrobot_das_controller}
to maintain a consistent gripper interface. We conduct 10 trials per task;
success requires completing the task while remaining upright and balanced.

\exphead{Runtime}
Across real-robot trials on an NVIDIA GeForce RTX 4090, mean inference
latencies are 64~ms for the DP and 104~ms for \textbf{WB-UMI}. Both are below
their respective update intervals of 100~ms and approximately 1.07~s.

\exphead{Drawer Closing}
Drawer closing tests coordinated reaching and body lowering. The robot
follows the EE plan to reach the handle and fully close the drawer, succeeding
in 9/10 trials. The closure miss in Fig.~\ref{fig:robot-demo-overview}
illustrates an embodiment gap: human demonstrators can reach the target by
crouching, whereas G1's shorter reach may require additional stepping or
leaning.

\exphead{Shelf PnP}
The robot transfers a bottle from the lower shelf to the upper shelf,
coordinating arm motion with changes in body height. Success requires stable
placement without dropping the bottle. The hierarchy succeeds in 8/10 trials,
demonstrating whole-body manipulation across different working heights.

\exphead{Ball Toss}
The robot must release a ball so that it lands in a target basket, testing
dynamic whole-body coordination. The hierarchy succeeds in 3/10 trials.
Failures were associated with gripper-release latency and DP prediction
errors during fast EE motion.

\exphead{Loco-PnP}
Loco-PnP combines locomotion and manipulation: the robot must walk to a table,
grasp a bottle, and place it at the target location without dropping it.
The hierarchy succeeds in 4/10 trials. The failures in
Fig.~\ref{fig:robot-demo-overview} suggest that execution drift can expose
the DP to out-of-distribution (OOD) observations, degrading subsequent EE
predictions. These predictions can, in turn, increase errors in motion
generation and execution. The left ankle roll curves in Fig.~\ref{fig:robot-demo-overview} show a gap
between the generated reference and SONIC's executed motion, illustrating
imperfect whole-body tracking on the real robot.

Taken together, these results answer \textbf{Q3}: the hierarchy transfers native UMI
skills without task-specific whole-body demonstrations, although reliability
remains lower for ball toss and Loco-PnP.

\section{Conclusion}

We presented \textbf{WB-UMI}, a task-agnostic model that generates real-time
whole-body motion from EE trajectories. By learning its motion prior
independently from mocap, \textbf{WB-UMI} connects a DP trained on native UMI
demonstrations to humanoid execution without body trackers or task-specific
whole-body demonstrations. On unseen action categories, EE look-ahead reduced
the position error from 5.77~cm to 4.25~cm and improved all other reported
generation metrics. The integrated system operated in real time and achieved
success rates of 90\%, 80\%, 30\%, and 40\% on drawer closing, shelf
pick-and-place, ball toss, and Loco-PnP, respectively. These results support
reliable transfer on the first two tasks and initial feasibility on the more
dynamic tasks.
Dynamic feasibility and error accumulation across EE planning, motion
generation, and control remain challenges. Future work will explore WBC data augmentation, and residual reinforcement
learning to improve execution. Scaling pretraining to more diverse UMI and
mocap datasets may further improve generalization across tasks and humanoid
platforms.

\section*{Acknowledgment}
This research was funded in part by the State Key Laboratory (SKL) of Biobased
Transportation Fuel Technology and the ZJU-YST Joint Research Center for
Fundamental Science. This work was also supported in part by the InnoHK
initiative of the Innovation and Technology Commission of the Hong Kong
Special Administrative Region Government via the Hong Kong Centre for
Logistics Robotics. The real-robot experiments were supported in part by
Mondo Robotics. We thank Jiexi Lyu and Yufeng Ji for insightful discussions.

\bibliographystyle{IEEEtran}
\bibliography{sections/references}

\end{document}